\documentclass[runningheads]{llncs}
\usepackage{comment}
\usepackage{graphicx}
\usepackage{amsmath}
\usepackage{booktabs}
\usepackage{multirow}
\usepackage[colorlinks=true, allcolors=blue]{hyperref}
\usepackage{overpic}
\usepackage{floatrow}

\begin{document}
\title{EFENet: Reference-based Video Super-Resolution with Enhanced Flow Estimation
}
\titlerunning{EFENet}
\authorrunning{Yaping Zhao, Mengqi Ji, Ruqi Huang, Bin Wang, Shengjin Wang}
\author{Yaping Zhao$^{1}$, Mengqi Ji$^{1}$, \thanks{Ruqi Huang is the corresponding author.}Ruqi Huang$^{1}$, Bin Wang$^{2}$, Shengjin Wang$^{1}$}
\institute{$^{1}$Tsinghua University, $^{2}$Hikvision}

\maketitle              

\begin{figure}[!h]
\vspace{-20pt}
    \centering
  \includegraphics[width=0.8\textwidth]{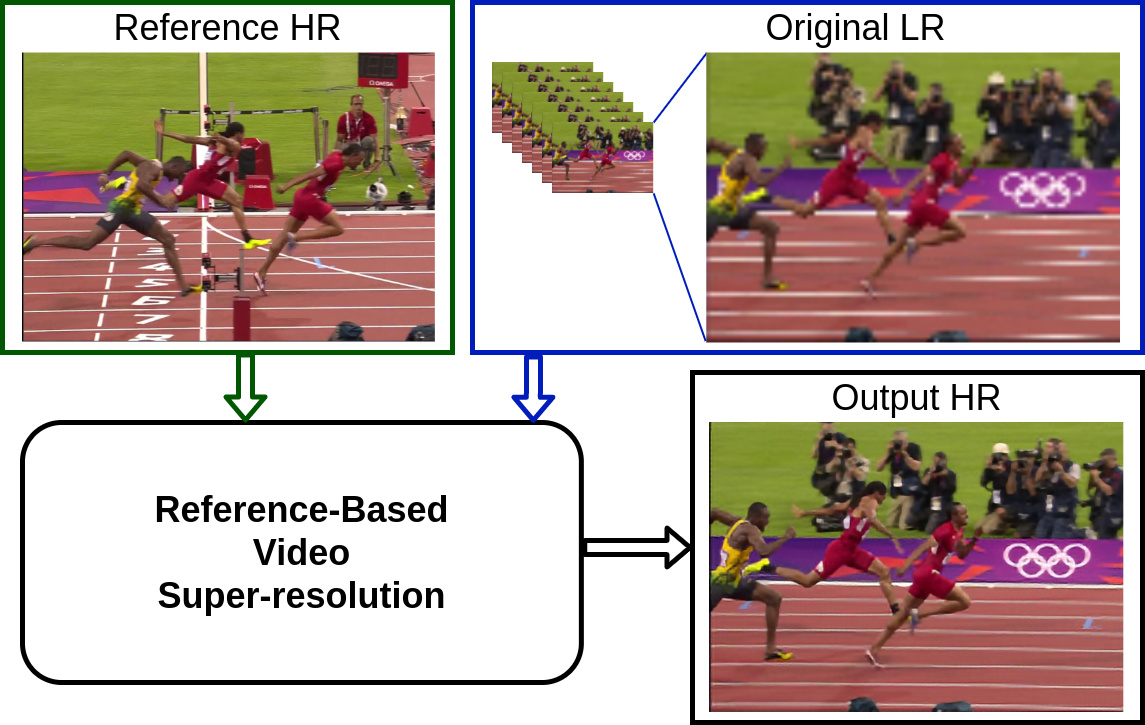}
  \vspace{-10pt}
  \caption{Given a high-resolution (HR) and multiple low-resolution (LR) frames as input, we propose a reference-based video super-resolution algorithm to output the HR frame.}
  \label{fig:teaser}
\end{figure}
 \vspace{-30pt}

\begin{abstract}
In this paper, we consider the problem of reference-based video super-resolution(RefVSR), i.e., how to utilize a high-resolution (HR) reference frame to super-resolve a low-resolution (LR) video sequence. The existing approaches to RefVSR essentially attempt to align the reference and the input sequence, in the presence of resolution gap and long temporal range. However, they either ignore temporal structure within the input sequence, or suffer accumulative alignment errors. To address these issues, we propose EFENet to exploit simultaneously the visual cues contained in the HR reference and the temporal information contained in the LR sequence. 
EFENet first globally estimates cross-scale flow between the reference and each LR frame. Then our novel flow refinement module of EFENet refines the flow regarding the furthest frame using all the estimated flows, which leverages the global temporal information within the sequence and therefore effectively reduces the alignment errors. We provide comprehensive evaluations to validate the strengths of our approach, and to demonstrate that the proposed framework outperforms the state-of-the-art methods.

\end{abstract}


\section{Introduction}
\label{sec:intro}
\vspace{-10pt}

In this paper, we consider the problem of reference-based video super-resolution (RefVSR), which is strongly motivated by the recent advances of hybrid camera systems. 
In general, our goal is to utilize a high-resolution (HR) reference frame to super-resolve a low-resolution (LR) video sequence captured by cameras at similar viewpoints (see, e.g., Fig.~\ref{fig:teaser}). 
While numerous methods~\cite{boominathan2014improving,refsr_zheng2017learning,zheng2018crossnet,tan2020crossnet++} have been proposed for reference-based super-resolution (RefSR), some of them even have been applied in practice such as giga-pixel imaging~\cite{brady2012multiscale,yuan2017multiscale,Zhang2020Multiscale-VR} and light-field reconstruction~\cite{boominathan2014improving,wang2016light}. The RefSR task for \textit{videos} remains challenging, due to the following two factors: 1) the large parallax and resolution gap (e.g., $ 4\times$) makes it difficult to transfer details from HR frame to LR ones; 2) the potential large temporal gap between the HR and LR frames can lead to significant viewpoint drift among frames, therefore makes the regarding correspondence estimation error-prone. 
To better align frames across different resolutions,
recent works~\cite{zheng2018crossnet,tan2020crossnet++} for RefSR apply a flow estimator from~\cite{FlowNet} to estimate the correspondence. Though such methods can be used in RefVSR by dividing the RefVSR task into multiple separated RefSR tasks, 
they discard the important temporal information contained in video sequence. Therefore, they have trouble handling long input sequence, as it is increasingly difficult to estimate correspondence between the reference and the LR frames as the temporal gap increases.
On the other hand, to exploit the temporal information from the input sequence, several works~\cite{video-super-resolution1,video-super-resolution2,videosr_tao2017detail} are proposed to leverage the temporal information between LR frames by flow estimation and motion compensation. However, they suffer the resolution discrepancy between the HR reference and the LR frames, hindering the usage of the rich visual cues from the reference.

To overcome the aforementioned difficulties, we propose an end-to-end neural network, EFENet
\footnote{ Code is available at \href{https://github.com/IndigoPurple/EFENet}{https://github.com/IndigoPurple/EFENet}.}
, for RefVSR. EFENet takes advantage of the state-of-the-art RefSR method~\textit{CrossNet++}~\cite{tan2020crossnet++}, but, more crucially, introduces a novel flow refinement module to improve the furthest frame alignment with \textit{all} the prior ones, allowing the usage of temporal information for long range video synthesis on top of the powerful correspondence estimation module. In the end, EFENet decodes the refined flow to output super-resolved video sequences. 
Note that, the existing video super-resolution methods~\cite{video-super-resolution1,video-super-resolution2,videosr_tao2017detail} locally estimate the correspondence between adjacent frames and propagate the information to the furthest frame accumulatively. In contrast to them, EFENet leverages the temporal information in a global manner, effectively reducing the accumulative alignment errors. 
We test our pipeline and several strong baselines on Vimeo90K and MPII datasets. Extensive experimental results show the substantial improvements of our method over the state-of-the-art methods.

Our contributions are listed as follows:
\begin{itemize}
    \item We propose an end-to-end network, EFENet, for RefVSR. By combining~\textit{CrossNet++} with a novel flow refinement module, EFENet effectively improves long sequence video synthesis quality for cross-scale camera systems.
    \item The flow refinement module of EFENet exploits globally the temporal information for flow enhancement, and therefore reduces alignment errors. 
    \item Extensive experiments demonstrate the superior performance of our method in comparison to the state-of-the-art methods.
\end{itemize}
\vspace{-10pt}
\section{Related Work}
\vspace{-5pt}
\subsection{Single-Image Super-Resolution}
\vspace{-5pt}
Since recovering the HR content from a single image is ill-posed, 
a large amount of image priors are proposed for single image super-resolution (SISR). 
Image priors proposed for the SR task include: gradient prior\cite{gradient_sun2008image}, sparsity prior\cite{sparse_yang2008image}, patch dictionary prior\cite{example_timofte2013anchored}, etc. 
With the development of deep learning,
Dong \emph{et al.} adapt a three-layers convolutional network~\cite{nnsr_dong2014learning} for SISR. 
Later, more advanced network structures are proposed to improve SISR results.
To further improve realism of SR outputs, 
Ledig \emph{et al.}~\cite{srgan_ledig2017photo} use a generative adversarial network to generate photo realistic SR result. 
Later, Sajjadi \emph{et al.} propose  EnhanceNet\cite{enhancenet_sajjadi2017enhancenet} to leverage  adversarial loss and perceptual loss to generate realistic textures.
However, SISR is limited to generate sharp image, especially for large scale factors, as it is essentially highly ill-posed.

\vspace{-5pt}
\subsection{Reference-based Super-Resolution}
\vspace{-5pt}
From the other perspective, the performance of SISR can be improved if an additional HR reference image similar to the LR image is given as input. 
Boominathan \emph{et al.}~\cite{refsr_boominathan2014improving} adopt high-resolution images captured by cameras from similar viewpoints as the reference, and propose a patch-based synthesis algorithm via non-local mean~\cite{buades2011non} for super-resolving the low-resolution light-field images.
Wu \emph{et al.}~\cite{wu2015novel} further improve it by employing patch registration before the nearest neighbor searching, then apply dictionary learning for reconstruction.
Wang \emph{et al.}~\cite{wang2016light} iterate the patch synthesizing step of~\cite{refsr_boominathan2014improving} for enriching the reference patch database. Zheng \emph{et al.}~\cite{zheng2017combining} decompose images into sub-bands by frequencies and apply patch matching for high-frequency sub-band reconstruction. In addition, Zheng \emph{et al.}~\cite{zheng2017combining} propose a cross-resolution patch matching and synthesis scheme for RefSR.
To improve RefSR, more recent works~\cite{zheng2018crossnet,tan2020crossnet++} utilize cross-scale optical flow networks and a warping-synthesis framework to perform RefSR prediction. Aslo there is work~\cite{SRNTT} leveraging patch-based correlation to perform non-rigid feature swapping for reference-based SR.

\vspace{-5pt}
\subsection{Video Super-Resolution}
\vspace{-5pt}
Compared to the image SR task, video SR enjoys extra cues given by the correlation between temporally adjacent frames. 
Essentially, video SR can be regarded as yet another temporal variant of reference-based SR task.
Recent methods~\cite{video-super-resolution1,video-super-resolution2,videosr_tao2017detail,Haris_2019_CVPR} leverage the temporal consistency between LR frames for video SR. Those works typically combine flow estimation and motion compensation to propagate the correspondence from previous time step to the current time step.
However, the existing works have difficulty leveraging the high-resolution visual information of additional reference frames. Moreover, the optical flow are estimated between low-resolution frames, resulting in degenerated correspondence.
\section{Method}



In this section, we start by giving a conceptual comparison among different RefVSR strategies (Sec.~\ref{ssec:pa}). And then we provide details of our pipeline, including the network structure (Sec.~\ref{ssec:ns}) and the optimization objective (Sec.~\ref{ssec:lf}).

\subsection{RefVSR Strategies Analysis}
\label{ssec:pa}
Given an HR reference $I^{H}_{t}$ at time step $t$ and an LR video sequence $\{I^{L}_{t+1},\cdots, I^{L}_{t+n}\}$, our goal is to generate an HR frame $\widehat{I}^{H}_{t+n}$ corresponding to the LR frame $I^{L}_{t+n}$ at time step $t+n$.

\begin{figure}[!h]
\vspace{-10pt}
    \centering
  \includegraphics[width=\textwidth]{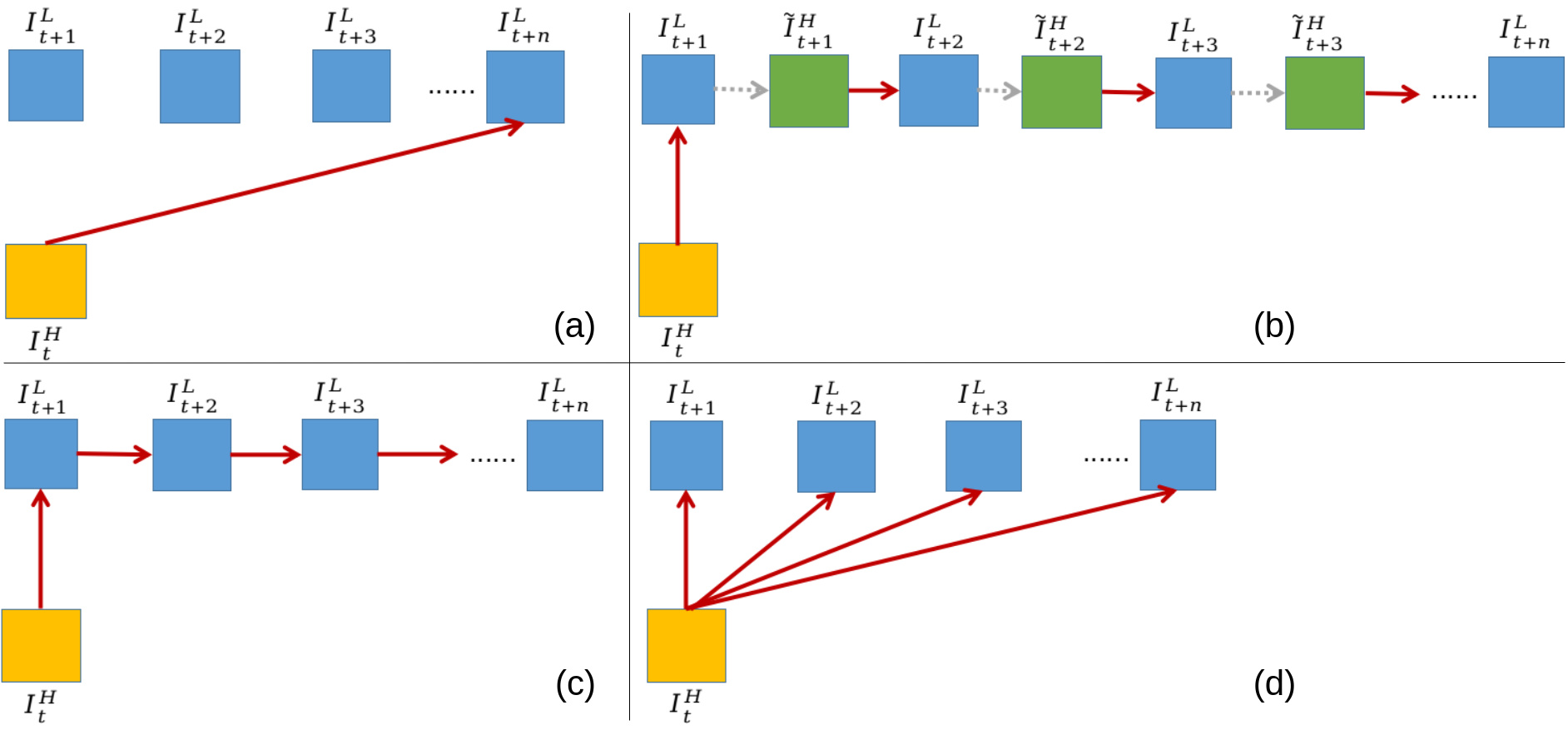}
  \caption{Different strategies for cross-scale correspondence estimation: (a) direct estimation~\cite{zheng2018crossnet,tan2020crossnet++}; (b) recursive warping following local correspondence estimation~\cite{video-super-resolution1,video-super-resolution2}; (c) sequential local correspondence composition~\cite{RBPN2019}; (d) refinement based on all global correspondence.}
  \label{fig:pre1}
\end{figure}

\vspace{-10pt}
A central issue of the RefVSR task is how to accurately estimate correspondence among frames that both across different resolutions and undergoing significant temporal (and thus viewpoint) drifts.
To see this, we first analyze different strategies taken in the prior arts, 
as shown in Fig.~\ref{fig:pre1}:
(a) each correspondence of the input frame pairs is estimated independently, limiting the ability to utilize temporal consistency~\cite{zheng2018crossnet,tan2020crossnet++};(b) the reference is processed and warped multiple times, resulting in increasing alignment errors~\cite{video-super-resolution1,video-super-resolution2}, and (c) correspondence is estimated on the adjacent LR frames~\cite{RBPN2019}, hindering the usage of the rich visual cues from the reference and resulting in degenerated correspondence. 

\begin{figure}[t!]
\vspace{-10pt}
    \centering
  \includegraphics[width=\textwidth]{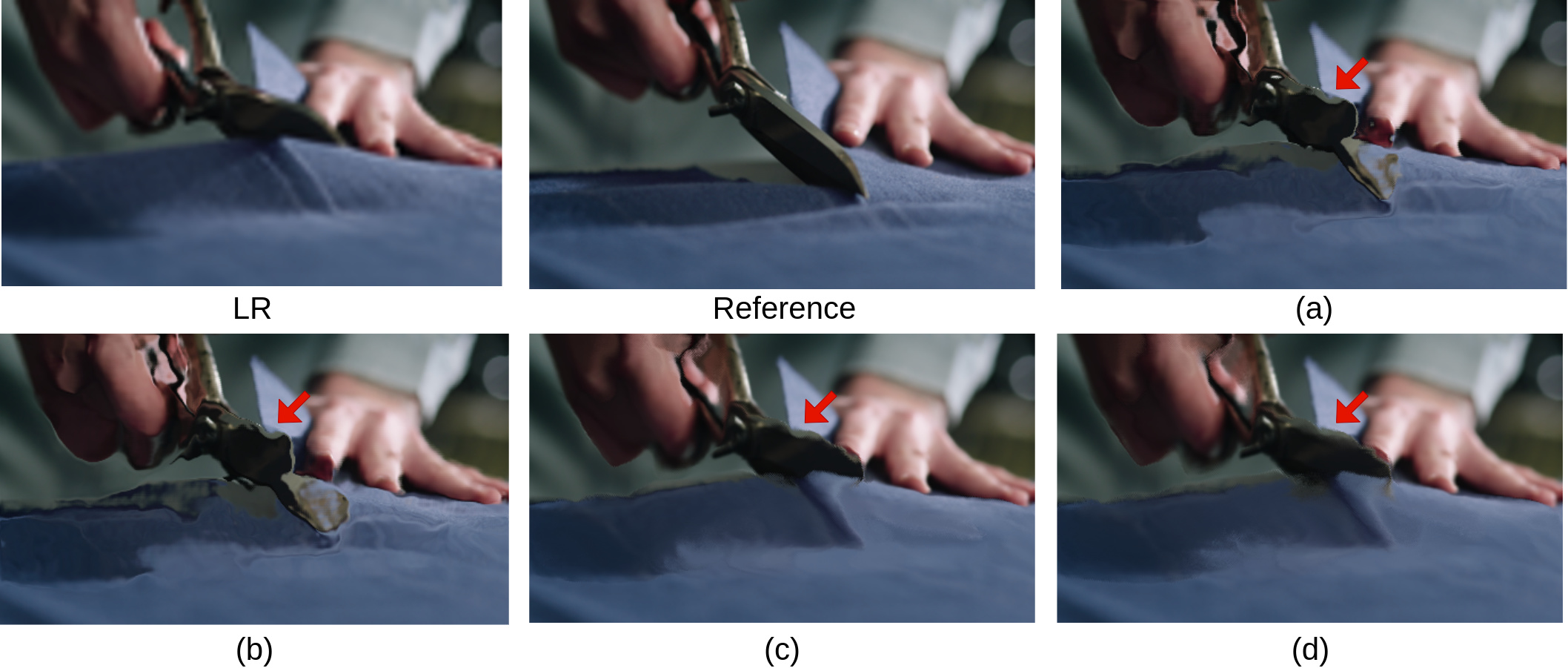}
  \caption{Given the LR frame and the reference, the image alignment results using different strategies: (a) direct estimation; (b) recursive warping; (c) sequential composition; (d) ours. Obviously, our strategy performs the best, see the discrepancies pointed out by the red arrows for comparison.}
  \vspace{-10pt}
  \label{fig:pre2}
\end{figure}

To overcome the limitations of above methods, we propose a new one shown in Fig.~\ref{fig:pre1} (d).
On one hand, different from (a), our method leverages temporal information to facilitate long range video synthesis.
On the other hand, different from (b) and (c), our method estimates cross-scale flow between the reference frame and \textit{each} of the LR frames, achieving full exploitation of visual cues from the reference. 
And then, we propose a flow refinement module that takes all the estimated flows as input and globally leverage the temporal information to reduce the alignment error.


To illustrate the effectiveness of our method, we re-implement the competing methods with the state-of-the-art flow estimator~\cite{FlowNet}. As shown in Fig.~\ref{fig:pre2}, our method (e) significantly improves the flow estimations method (a) of \textit{CrossNet++}~\cite{tan2020crossnet++} by taking temporal information into consideration. Moreover, the way we leverage such information is superior to the naive combination of \textit{CrossNet++}~\cite{tan2020crossnet++} and the existing VSR methods~\cite{video-super-resolution1,video-super-resolution2,RBPN2019} (b, c).

\vspace{-5pt}
\subsection{Network Structure}
\label{ssec:ns}

\begin{figure}[!h]
\vspace{-10pt}
    \centering
  \includegraphics[width=\textwidth]{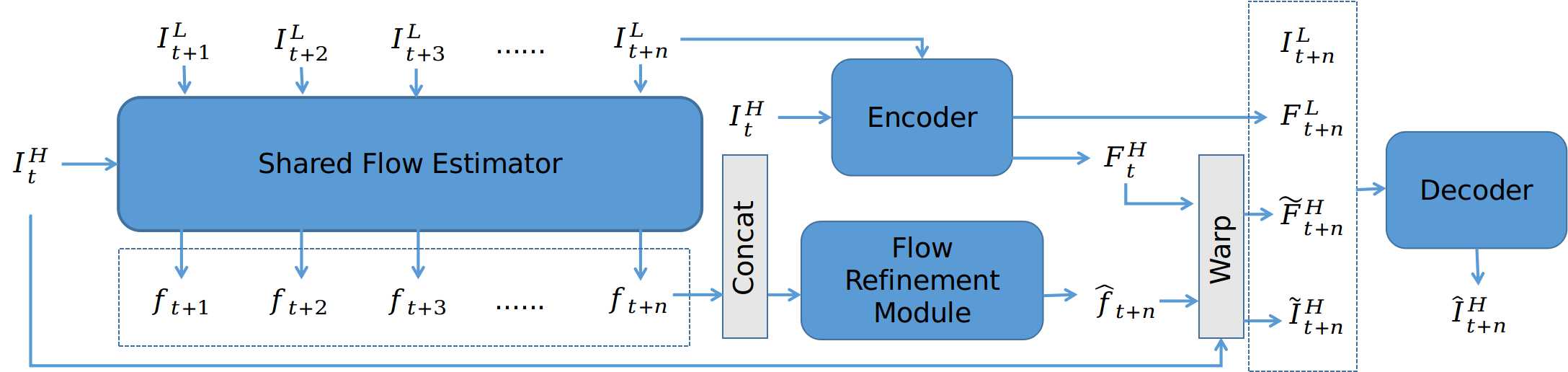}
  \caption{Our deep neural network for RefVSR. It contains a shared flow estimator (left), a flow refinement module (middle), and an encoder-decoder synthesis network (right).}
  \vspace{-5pt}
  \label{fig:vsr-pipeline}
\end{figure}

Fig.~\ref{fig:vsr-pipeline} illustrates the architecture of EFENet, which contains a flow estimator, a flow refinement module, and a synthesis network. 

\vspace{-25pt}

\vspace{-5pt}
\subsubsection{Shared Flow Estimator Module}
\label{sss:1}
As shown on the left of Fig.~\ref{fig:vsr-pipeline}, we first estimate the correspondence between the reference frame $I^{H}_{t}$ and each of LR frames $\{I^{L}_{t+1},\cdots, I^{L}_{t+n}\}$, resulting $n$ flow maps $\{f_{t+1},\cdots, f_{t+n}\}$. Our flow estimator is based on a recently proposed the cross-scale flow estimator~\cite{tan2020crossnet++}. 
Formally,
\begin{align}
\begin{aligned}
\label{eq:flow}
f_{t+i} = {\mbox{Net}_{\mbox{flow}}}(I^H_t, {I^L_{t+i}}^\uparrow), i = 0, \cdots, n
\end{aligned}
\end{align}
where $\mbox{Net}_{\mbox{flow}}$ denotes the flow estimator, ${I^L_{t+i}}^\uparrow$ denotes the bicubicly upsampled LR frame and $f_{t+i}$ denotes the output flow.

\vspace{-10pt}
\subsubsection{Flow Refinement Module}
\label{sss:2}

Obviously, estimating flow for each frame independently ignores the temporal relationships among them, thus leads to sub-optimal results in the presence of large motion. 
To address this issue, we propose a flow refinement module which takes
the concatenation of 
the generated flow maps at multiple time steps as input and output the refined flow at time step $t+n$. 

In order to globally leverage prior flows to refine the temporally furthest flow, we adopt a U-Net~\cite{U-Net} like structure with 4 strided convolution and de-convolution layers as an encoder-decoder. In fact, using such an encoder-decoder have two advantages: on one hand, the encoder can capture high-level information, including semantic and temporal information, in the hierarchical convolution process; on the other hand, the decoder, which uses skip-connections in the de-convolution process, integrates features at different scales to ensure the enhancement of the reconstructed flow.

Specifically, we let
\begin{align}
\begin{aligned}
\label{eq:flow}
\hat{f}_{t+n} = {\mbox{Net}_{\mbox{refine}}}([f_{t+1},\cdots,f_{t+n}]),
\end{aligned}
\end{align}
where $[\cdot,\cdot]$ represents concatenation operation, $\mbox{Net}_{\mbox{refine}}$ represents the refinement module and $\hat{f}_{t+n}$ is the refined optical flow.

\vspace{-10pt}
\subsubsection{Encoder-Decoder Synthesis Network}
\label{sss:3}
After the flow refinement, we use another U-Net like encoder-decoder with 4 strided convolution and de-convolution layers to synthesize the SR result.
Specifically, we first apply an encoder $\mbox{Net}_E$ 
to extract the feature map at scale 1 from ${I^L_{t+n}}^\uparrow$ and $I^H_{t}$, and repeatedly convolve the feature map at the scale $i - 1$ (for $1 < i \leq 4$) to extract the feature map at scale i:

\begin{align}
\begin{aligned}
\{F^L_{t+n, s} \} &= \mbox{Net}_E({I^L_{t+n}}^\uparrow), \\
\{F^H_{t, s}\} &= \mbox{Net}_E(I^H_{t}), s = 1, 2, 3, 4,
\end{aligned}
\end{align}
where $F^L_{t+n,s} $ is the feature map of up-sampled LR frame ${I^L_{t+n, s}}^{\uparrow}$ at scale $s$, and $F^H_{t,s}$ is the feature map of reference image $I^H_{t}$ at scale $s$.

After the feature encoding, we leverage the refined flow $\hat{f}_{t+n}$ from the previous step to warp the reference image $I^{H}_t$ and reference feature maps $\{F^H_{t,s}\}$:
\begin{align}
\begin{aligned}
\label{eq:warp}
\widetilde{I}^{H}_{t+n} &= \mathcal{W}(I^{H}_t, \hat{f}_{t+n}),\\
\{\widetilde{F}^{H}_{t+n, s}\} &= \mathcal{W}(\{F^{H}_{t, s}\}, \hat{f}_{t+n}),
s = 1,2,3,4,
\end{aligned}
\end{align}
where $\mathcal{W}(\cdot,f)$ denotes backward warping operation that warps an image or feature maps according to the flow $f$.

Finally, we design a
decoder ${\mbox{Net}_{D}}$ to generate the final output. Specifically, the feature maps $\{F^L_{t+n, s}\}$, upsampled LR frame ${I^L_{t+n}}^\uparrow$, warped reference features $\{\widetilde{F}^{H}_{t+n, s}\}$ and warped reference $\widetilde{I}^{H}_{t+n}$ are concatenated
to generate $\widehat{I}^{H}_{t+n}$:

\begin{align}
\label{eq:vsr-result}
\hat{I}^{H}_{t+n} = {\mbox{Net}_{D}}([{I^{L}_{t+n}}^{\uparrow}, \widetilde{I}^{H}_{t+n},\{F^L_{t+n,s}\}, \{\widetilde{F}^{H}_{t+n, s}\}]),
s = 1, 2, 3, 4.
\end{align}

\subsection{Loss Function}
\label{ssec:lf}
To train the network, we employ a combination of reconstruction loss and a set of warping loss. Specifically, let $\{I^{H}_{t+1},\cdots, I^{H}_{t+n}\}$ denote the HR ground-truth (GT) of LR frames $\{I^{L}_{t+1},\cdots, I^{L}_{t+n}\}$, we first formulate a reconstruction loss:
\begin{equation}
\mathcal{L}_{SR} =  \frac{1}{N}\sum_{j=1}^N\sum_k\rho(\hat{I}^{H}_{t+n} - I^{H}_{t+n}),
\end{equation}
where $\rho(x) = \sqrt{x^2 + 0.001^2}$ is the Charbonnier penalty function, $N$ is the number of samples, $j$ and $k$ iterate over samples and spatial locations respectively.

In addition, to regularize the estimated flow at each time step, we impose constraint on the estimated optical flows $f_{t+1},\cdots,f_{t+n}$. Specifically,
\begin{equation}
    \mathcal{L}_{\mbox{flow1}} = \frac{1}{2N}\sum_{i=1}^n\sum_{j=1}^N\sum_k{||\mathcal{W}(I^{H}_{t}, {f}_{t+i}) - I^{H}_{t+i})||}_2^2,
\end{equation}
where $\mathcal{W}(I^{H}_{t+i}, {f}_{t+i})$ denotes backward version of $I^{H}_{t+i}$ using flow ${f}_{t+i}$; $n$ is the number of time steps, $N$ is the number of samples, $i$, $j$ and $k$ iterate over time steps, training samples and spatial locations respectively.

Similarly, we impose constraint on the enhanced flow $\hat{f}_{t+n}$:
\begin{equation}
    \mathcal{L}_{\mbox{flow2}} = ||\mathcal{W}(I^{H}_{t}, \hat{f}_{t+n}) - I^{H}_{t+n}||_2^2.
\end{equation}
\section{Experiment}

\subsection{Evaluation}

We evaluate our approach on the large-scale and representative datasets including Vimeo90K~\cite{xue17toflow} and MPII~\cite{andriluka14cvpr}. Vimeo90K contains 89,800 video clips with 7 frames, while most sequences of MPII have about 40 frames. 
We take the initial frame from each video clip as the reference, and down-sample the remaining frames as the LR inputs.
Following~\cite{zheng2018crossnet,tan2020crossnet++}, we set the resolution gap as $4\times$. 
Finally, to quantitatively evaluate the results, we use image quality metrics PSNR and SSIM~\cite{wang2004image}.

We train our network for $150,000$ iterations on Vimeo90K, and apply the Adam~\cite{ADAM} with $\beta_1=0.9, \beta_2=0.999$ as the optimizer.
For comparison, we test RefSR methods including \textit{CrossNet}~\cite{zheng2018crossnet}, \textit{CrossNet++}~\cite{tan2020crossnet++}, \textit{SRNTT}~\cite{SRNTT} and VSR methods including
\textit{ToFlow}~\cite{xue17toflow}, \textit{MEMC-Net}~\cite{MEMC-Net}, \textit{RBPN}~\cite{Haris_2019_CVPR}, and \textit{EDVR}~\cite{wang2019edvr}. In addition, we train a variant of RBPN~\cite{RBPN2019} such that its input contains the HR reference, resulting a method called \textit{RBPN w/ ref}.

\begin{figure}[H]
  \includegraphics[width=\textwidth]{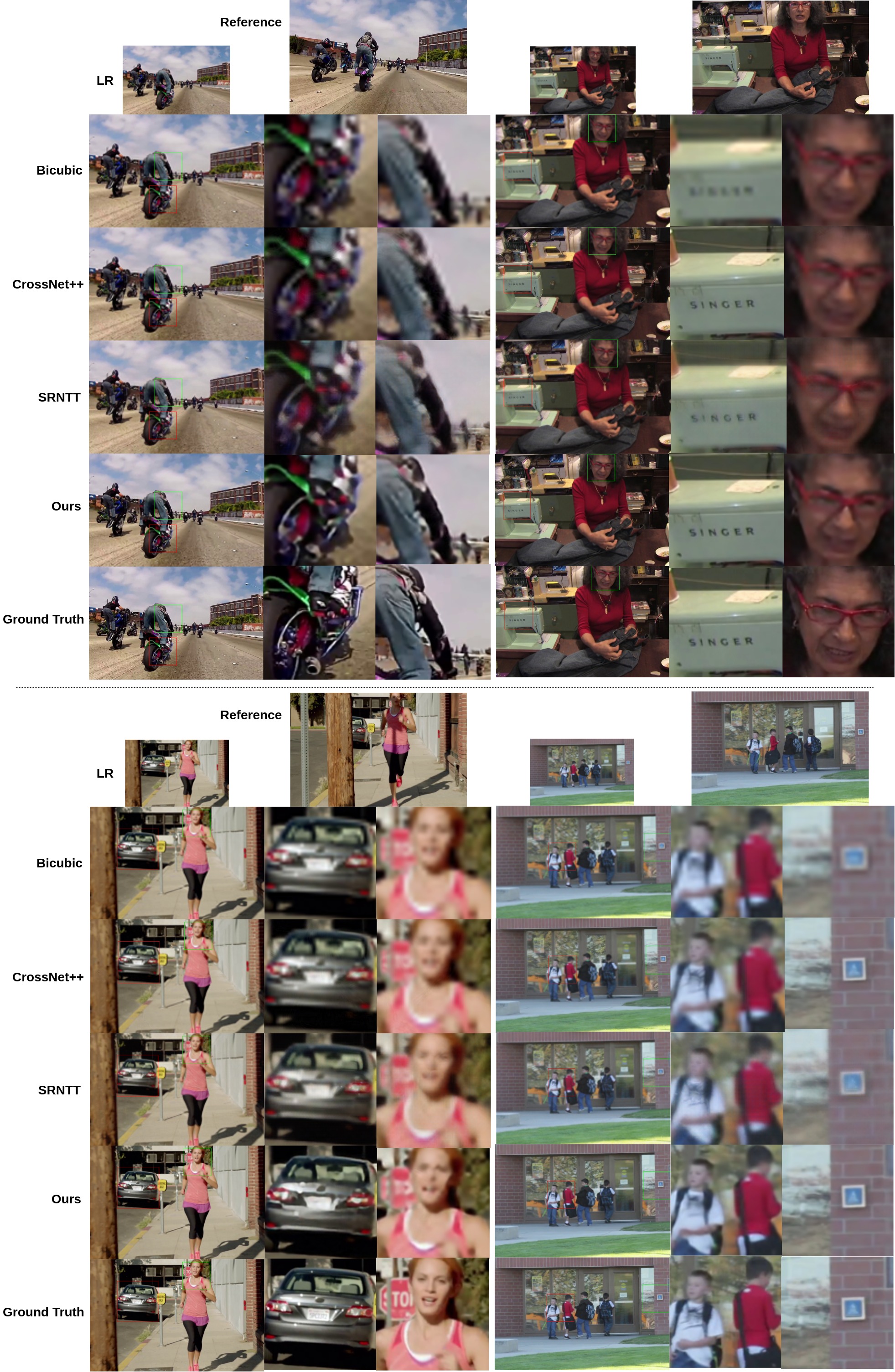}
  \caption{ Comparisons on MPII dataset under the cross-scale $4\times$ settings.}
  \label{fig:vsr-mpii1}
\end{figure}

\begin{figure}[H]
  \includegraphics[width=\textwidth]{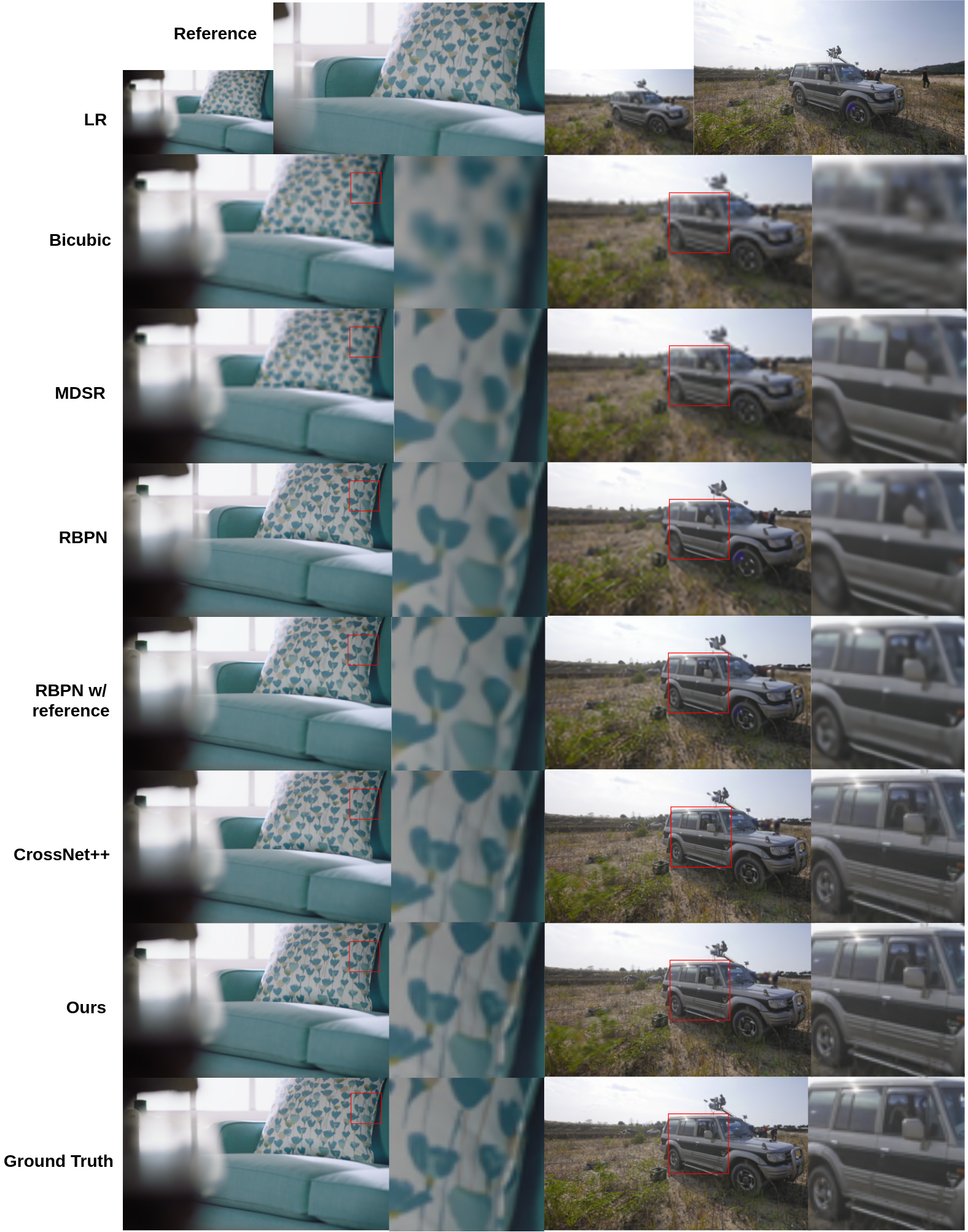}
  \caption{ Comparisons on Vimeo90K dataset under the cross-scale $4\times$ settings.}
  \label{fig:vsr-vimeo}
\end{figure}

\vspace{-20pt}

From Table \ref{tab:tb2}, it is evident that our method outperforms all the baselines. 
Moreover, we observe that: 
i) our method outperforms the existing reference-based method as we leverage the temporal information across frames to estimate better correspondence;
ii) our method outperforms the RefVSR approach \textit{RBPN w/ ref} which leverages sequential local correspondence composition, showing the advantage of our proposed flow estimation module.

We also provide qualitative comparisons in Fig.~\ref{fig:vsr-mpii1},~\ref{fig:vsr-vimeo}.
From the figures, we observe that: 
i) among the reference-based approaches including \textit{CrossNet}, \textit{CrossNet++}, \textit{SRNTT} and \textit{RBPN w/ ref}, our method generates the most sharp and visually plausible images;
ii) while \textit{CrossNet++} achieves good performance on super-resolving static objects (e.g., text, buildings), it fails in non-rigid deformation cases (e.g., human body, face).

\begin{table*}
 \caption{Left: quantitative evaluations of existing SR methods on Vimeo90K dataset. Right: quantitative evaluations of ablation methods on Vimeo90K and MPII datasets.}
\begin{floatrow}
\begin{tabular}{ccccc}
\hline
 Inputs  & Methods  & PSNR   & SSIM \\
\hline
 1 LR  & Bicubic  & $29.79$  & $0.90$   \\
\hline
 \multirow{3}*{7 LR} & TOFlow~\cite{xue17toflow}& $33.08$ & $0.94$ \\
& MEMC-Net~\cite{MEMC-Net} & $33.47$ & $0.95$ \\
& RBPN~\cite{RBPN2019} & $35.32$ & $0.95$ \\
& EDVR~\cite{wang2019edvr} & $35.79$ & $0.94$ \\
\hline
 \multirow{2}*{1 LR+1 HR} & CrossNet~\cite{zheng2018crossnet} & $36.71$ & $0.95$ \\
& CrossNet++~\cite{tan2020crossnet++} & $38.95$ & $\textbf{0.97}$ \\
\hline
 \multirow{2}*{ 7 LR+1 HR} & RBPN w/ ref & $36.60$ & $0.95$ \\
& \textbf{ours} & $\textbf{39.25}$ & $\textbf{0.97}$ \\
\bottomrule
\end{tabular}
\hfill
\hspace{5pt}
\begin{tabular}{cccc}
\hline
Dataset  & Methods  & PSNR   & SSIM \\
\hline
\multirow{4}*{Vimeo} & CrossNet++~\cite{tan2020crossnet++}   & $38.95$ & $0.97$\\
& Setup1  & $38.85$ & $0.97$ \\
& Setup2 & $39.14$ & $0.97$ \\
& \textbf{ours} & $\textbf{39.25}$ & $0.97$ \\
\hline
\multirow{4}*{MPII} & CrossNet++ ~\cite{tan2020crossnet++}   & $30.19$ & $0.87$\\
& Setup1  & $30.11$ & $0.87$ \\
& Setup2 & $30.60$ & $0.88$ \\
& \textbf{ours} & $\textbf{30.76}$ & $\textbf{0.89}$ \\
\bottomrule
\end{tabular}

 \label{tab:tb2}
\end{floatrow}
\end{table*}

\vspace{-10pt}
\subsection{Ablation Study}
To compare different correspondence estimation designs, we propose a set of ablation experiments. Specifically, similar to Fig.~\ref{fig:pre1} (a), we use the original network of \textit{CrossNet++} for synthesis.
We also implemented alignment modules based on the diagram of (b) and (c) in Fig.~\ref{fig:pre1}, resulting two variants of \textit{CrossNet++} called \textit{Setup 1} and \textit{Setup 2}. From Table \ref{tab:tb2}, we observe that:
i) when facing long temporal ranges on MPII dataset, the performance of \textit{CrossNet++} decayed more than ours;
ii) naively combining \textit{CrossNet++} with existing VSR strategies performs worse than our method, and even causes degeneration to original \textit{CrossNet++}.



\section{Conclusion}
We present an end-to-end network EFENet for RefVSR, which globally estimates cross-scale flow between the reference frame and each LR frame, and then improves the flow estimation with a novel flow refinement module. 
The core of EFENet is a novel flow refinement module, which exploit simultaneously the visual cues contained in the HR reference and the temporal information contained in the LR sequence to boost the super-resolution performance. 
Finally, we provide comprehensive evaluations and comparisons with previous methods to validate the strengths of our approach and demonstrate that the proposed network is able to outperform the state-of-the-art methods.


\end{document}